\documentclass[12pt,onecolumn,draftclsnofoot]{IEEEtran}

\IEEEoverridecommandlockouts     
\usepackage[letterpaper,
            top=1in,
            bottom=1in,
            left=1in,
            right=1in]{geometry}
\usepackage{cite}
\usepackage{amsmath,amssymb,amsfonts}
\usepackage{booktabs}
\usepackage{algorithmic}
\usepackage{graphicx}
\usepackage{textcomp}
\usepackage{booktabs}
\usepackage{multirow} 
\usepackage{xcolor}
\def\BibTeX{{\rm B\kern-.05em{\sc i\kern-.025em b}\kern-.08em
    T\kern-.1667em\lower.7ex\hbox{E}\kern-.125emX}}
    
\begin{document}

\title{\vspace*{.8cm}
\normalfont\fontsize{16}{18}\selectfont\bfseries Trustworthiness of Stochastic Gradient Descent in Distributed Learning}

\author{
\IEEEauthorblockN{
Hongyang Li\IEEEauthorrefmark{1}, 
Caesar Wu\IEEEauthorrefmark{1}, 
Mohammed Chadli\IEEEauthorrefmark{2},\\
Said Mammar\IEEEauthorrefmark{2}, 
Pascal Bouvry\IEEEauthorrefmark{1}
}

\IEEEauthorblockA{\IEEEauthorrefmark{1}SnT, University of Luxembourg, Luxembourg\\
\texttt{\{hongyang.li, caesar.wu, pascal.bouvry\}@uni.lu}}

\IEEEauthorblockA{\IEEEauthorrefmark{2}Paris-Saclay University, France\\
\texttt{\{mohammed.chadli, said.mammar\}@univ-evry.fr}}
}

\maketitle

\begin{abstract}

Distributed learning (DL) uses multiple nodes to accelerate training, enabling efficient optimization of large-scale models. Stochastic Gradient Descent (SGD), a key optimization algorithm, plays a central role in this process. However, communication bottlenecks often limit scalability and efficiency, leading to increasing adoption of compressed SGD techniques to alleviate these challenges. Despite addressing communication overheads, compressed SGD introduces trustworthiness concerns, as gradient exchanges among nodes are vulnerable to attacks like gradient inversion (GradInv) and membership inference attacks (MIA). The trustworthiness of compressed SGD remains unexplored, leaving important questions about its reliability unanswered. 

In this paper, we provide a trustworthiness evaluation of compressed versus uncompressed SGD. Specifically, we conducted empirical studies using GradInv attacks, revealing that compressed SGD demonstrates significantly higher resistance to privacy leakage compared to uncompressed SGD. In addition, our findings suggest that MIA may not be a reliable metric for assessing privacy risks in distributed learning. 

\end{abstract}


\vspace{-3mm}

\section{Introduction}
\label{sec:intro}

DL is the method used to accelerate the training of deep learning models by distributing training tasks to multiple computing nodes \cite{mcdonald2009efficient}. However, as data scales continue to grow, the complexity of model gradients increases accordingly, for example, consider deep learning training on ImageNet \cite{deng2009imagenet}, which contains more than 14 million labeled images and topics with approximately 22,000 categories, which leads to constraints on communication efficiency \cite{beznosikov2023biased}. 

Gradient compression aims to reduce communication overhead during gradient transmission between multiple nodes, which improves the computational efficiency of the system \cite{wang2018atomo, lin2017deep, 2019powersgd}, thus this has emerged as an effective optimization technique in distributed learning, especially when training complex models to process large-scale data. Among various gradient compression techniques, PowerSGD \cite{2019powersgd} and Top-K SGD \cite{alistarh2018convergence} have emerged as prominent solutions due to their ability to substantially reduce communication costs while preserving scalability and model accuracy in distributed large-scale learning. These two algorithms are particularly suitable for our study as they represent fundamental approaches to gradient compression: PowerSGD uses a low-rank approximation, while Top-K SGD implements gradient sparsification by retaining only the top K elements (by absolute value) of the gradient. Both techniques are widely recognized for their practical effectiveness, especially when combined, to varying extents, with advanced features such as error feedback, warm start, all-reduce, making them ideal candidates of compressed SGD for assessing privacy risks in distributed deep learning systems. 

Although distributed deep learning systems share model gradients instead of raw data, they are still vulnerable to indirect privacy attacks \cite{zhu2019deep}. One notable attack is GradInv, where an adversary attempts to reconstruct original training data or extract sensitive information from the shared gradients during training. GradInv has been demonstrated as an effective method for exposing private data, making it a significant privacy concern. Given this context, gradient reconstruction attacks are considered a powerful technique for evaluating the robustness of distributed algorithms. Another relevant attack is MIA \cite{shokri2017membership, song2021systematic, yeom2018privacy}, where an attacker infers whether a specific data point was part of the model’s original training. Although previous studies, such as \cite{zhu2019deep, huang2021evaluating, wei2020framework}, have examined sparsified gradient techniques as potential defenses against passive attacks, their focus has been limited to sparsification alone under single attackers GradInv, without a comprehensive analysis of a wider range of compressed SGD algorithms\cite{zhu2019deep, yin2021see}.

To address this gap, we conducted an in-depth comprehensive evaluation with two attackers for two representative methods: PowerSGD, which leverages low-rank approximation, and the widely-used Top-K SGD, which employs threshold-based sparsification. 

This study aims to offer a theoretical analysis alongside empirical validation, providing insights into the robustness of these techniques in distributed learning settings, and provides an empirical evaluation of the trustworthiness of compressed SGD algorithms in distributed learning. We assess their performance across various datasets, compression levels, and compare them with uncompressed SGD. Our results show that compressed SGD offers stronger resistance to gradient inversion attacks, indicating better privacy protection. We also find that MIAs show low sensitivity to both compressed and uncompressed SGD, suggesting MIA may not be an ideal privacy metric for these algorithms.

\section{Related Work and Problem Setup}
\label{sec:rela}
\subsection{Gradient Compression}

Several strategies have been proposed to address communication bottleneck problem in distributed deep learning. Quantization-based methods have gained attention for their simplicity and effectiveness, includes Quantized SGD (QSGD)\cite{alistarh2017qsgd, gandikota2022vqsgd} which employs lossy compression by quantizing gradients before transmission between nodes. signSGD where only exchange the sign of each gradient vector \cite{bernstein2018signsgd}, although this effectiveness is evident, it can have limited generalization capacity. To advance this, EF-SGD compensates with Error Feedback, enhancing the robustness of signSGD \cite{karimireddy2019error}; Another line of work focuses on sparsification, by transmitting only the most critical parts of the gradients, \cite{aji2017sparse} introduced a sparse SGD technique that sets 99\% of small gradient updates to zero, transmitting only the sparse matrix, and more compressed methods of sparsification proposed by \cite{lin2017deep, stich2018sparsified, alistarh2018convergence} ; Considering the scalability challenges and the high computational cost of compression methods in large-scale federated learning, low-rank compression method like PowerSGD and Top-K SGD stands out as the excellent compression SGD algorithms \cite{wangni2018gradient, 2019powersgd}.

\subsection{Top-K SGD and PowerSGD}

\textbf{Top-K SGD} reduces the communication load by selecting only the top $K$ gradient elements with the largest magnitudes, forming a sparse representation. This approach maintains a close approximation to the full gradient, allowing the transmission of the most important information while discarding less critical components \cite{wangni2018gradient}. The addition of error feedback further enhances Top-K SGD by accumulating any missed gradient information in subsequent updates. This algorithm is widely used in large-scale distributed learning due to its simplicity and effectiveness in reducing the volume of data exchanged with the server \cite{2019powersgd}.

\textbf{PowerSGD} offers an alternative approach is to compress the gradients by approximating them with low-rank matrices. Instead of transmitting the full gradient vector, achieving substantial communication reduction in structured layers, and significantly reducing the communication cost. Vogels et all. demonstrated that PowerSGD surpasses standard SGD in performance on a multiple GPUs setup, even on high-speed networks. By significantly reducing communication time 54\% for ResNet18 on CIFAR10, also lowered overall training time by 24\%, all while maintaining model accuracy within 1\% of uncompressed SGD \cite{2019powersgd}. These make PowerSGD a superior choice compared to other compression-based SGD algorithms.

\subsection{Passive Attackers: GradInv and MIA}
In distributed learning, gradient inversion is a widely used attack technique that exploits gradient updates to infer model inputs. Prior works, such as \cite{aono2017privacy} and \cite{melis2019exploiting}, demonstrated that sharing gradients in shallow neural networks can lead to significant leakage of original information. Building on this, \cite{zhu2019deep, geiping2020inverting} extended these techniques, showing that even in deep networks, shared gradients can effectively recover original image data. These studies underscore the privacy risks posed by gradient sharing during deep learning model training.  This work benchmarks at the state-of-the-art cosine similarity gradient inversion algorithm from \cite{geiping2020inverting} to evaluate the trustworthy performance of compressed SGD methods in DL with ResNet-18 \cite{he2016deep}.

Moreover, we select three common MIA methods—based on prediction confidence, loss values, and cross-entropy to evaluate the privacy risks of machine learning models. The prediction-based MIA method analyzes the model's output probabilities, asserting that models typically have higher prediction confidence for training data samples \cite{shokri2017membership}, the loss-based MIA method calculates the loss value for each sample, assuming that training data samples generally have lower losses \cite{yeom2018privacy}, and the cross-entropy-based MIA method assesses the cross-entropy between the model's predicted distribution and the true labels; lower cross-entropy values indicate a better fit of the model to the sample, making it more likely to belong to the training set \cite{song2021systematic}. 

\subsection{Distributed Learning Setting}

We consider a distributed learning setup with $N$ clients, each having a local dataset $\mathcal{D}_w$ for $w = 1, 2, \ldots, N$. The objective is to train a global model with parameters \(\theta \in \mathbb{R}^d\) by minimizing the average loss:
\begin{equation}
    \min_{\theta} \frac{1}{N} \sum_{w=1}^N \mathbb{E}_{(x, y) \sim \mathcal{D}_w} [\mathcal{L}(x, y; \theta)],
\end{equation}
where \(\mathcal{L}(x, y; \theta)\) is the loss function computed for the model parameters \(\theta\), input \(x\), and corresponding label \(y\) from the local dataset \(\mathcal{D}_w\) \cite{mcmahan2017communication, mcmahan2016federated}.

\subsection{GradInv Description}

The objective of the GradInv\cite{geiping2020inverting} method is to minimize \(\mathcal{L}_{\text{recon}}(x)\), such that the generated gradient \(\nabla_{\theta} \mathcal{L}(x, y; \theta)\) aligns as closely as possible with the:
\begin{equation}
\min_{x} \mathcal{L}_{\text{recon}}(x) = 1 - \frac{\langle \nabla_{\theta} \mathcal{L}(x, y; \theta), g \rangle}{\|\nabla_{\theta} \mathcal{L}(x, y; \theta)\| \|g\|}
\end{equation}
where \(\nabla_{\theta} \mathcal{L}(x, y; \theta)\) is the gradient of the loss with respect to model parameters \(\theta\) for input \(x\), and \(y\) is the corresponding label (in some cases the label may be assumed to be known), and \(g\) is the given gradient, the adversary's goal is to minimize $\mathcal{L}_{\text{recon}}(x)$ such that it approaches 0.

\section{Theoretical Analysis}
\label{sec:theo}
Having outlined the attack mechanisms and learning setup, we next provide theoretical insights into why these gradient compression methods may hinder gradient inversion. Here we provide a theoretical proof that demonstrates why gradient inversion attacks in PowerSGD and Top-K SGD fail to converge. 

\subsection{Convergence Analysis}

\textbf{PowerSGD} approximates a full-rank gradient matrix \( M \in \mathbb{R}^{n \times m} \) by a rank-\( r \) matrix \( \hat{M} = P Q^\top \), where \( P \in \mathbb{R}^{n \times r} \), \( Q \in \mathbb{R}^{m \times r} \), and \( r \ll \min(n, m) \). This low-rank representation significantly reduces communication overhead in distributed training by transmitting only the compact factors \( P \) and \( Q \), instead of the full gradient matrix \( M \).

PowerSGD constructs the approximation without explicitly computing the truncated singular value decomposition (SVD). Instead, it employs a randomized projection mechanism to estimate the dominant subspace of \( M \). 

To understand the quality of this approximation, we compare it against the optimal rank-\( r \) matrix given by truncated SVD. Any matrix \( M \in \mathbb{R}^{n \times m} \) can be decomposed as:
\begin{equation}
M = \sum_{i=1}^{r_{\text{full}}} \sigma_i u_i v_i^\top,
\label{eq:svd}
\end{equation}
where \( \sigma_i \) are the singular values in descending order, and \( u_i \in \mathbb{R}^n \), \( v_i \in \mathbb{R}^m \) are the corresponding left and right singular vectors.

The best rank-\( r \) approximation of \( M \) in the Frobenius norm sense is the truncated SVD:
\begin{equation}
\hat{M}_{\text{SVD}} = \sum_{i=1}^{r} \sigma_i u_i v_i^\top,
\label{eq:truncated_svd}
\end{equation}
and the approximation error is given by:
\begin{equation}
\|M - \hat{M}_{\text{SVD}}\|_F = \sqrt{\sum_{i=r+1}^{r_{\text{full}}} \sigma_i^2}.
\label{eq:frobenius_error}
\end{equation}

Although PowerSGD does not compute this optimal \(\hat{M}_{\text{SVD}}\), its approximation \(\hat{M} = P Q^\top\) is designed to capture a similar low-rank structure. Under mild assumptions---such as rapid singular value decay and sufficient orthogonality of projections---the error of PowerSGD's approximation can be bounded above by the error in Equation~\eqref{eq:frobenius_error}. Therefore, we treat the truncated SVD error as a theoretical benchmark for analyzing the information loss caused by PowerSGD.

This approximation error directly limits the quality of the compressed gradients, which in turn degrades the effectiveness of gradient inversion attacks, as significant information components may be irrecoverably lost in the low-rank compression.

\textbf{Top-K SGD} compresses each gradient vector \( \mathbf{g}_i \in \mathbb{R}^d \) by retaining only the \( K \) entries with the largest magnitudes (in terms of absolute value), and setting the remaining entries to zero. This results in a sparse vector \( \hat{\mathbf{g}}_i \in \mathbb{R}^d \), where \cite{wangni2018gradient}:
\[
\hat{g}_{i,j} = 
\begin{cases}
g_{i,j}, & \text{if } j \in \mathcal{I}_i, \\
0, & \text{otherwise},
\end{cases}
\]
and \( \mathcal{I}_i \subset \{1, \dots, d\} \) denotes the indices of the top \( K \) elements in absolute value.

The approximation error introduced by this hard thresholding operation can be quantified by the Euclidean norm of the discarded entries:
\begin{equation}
\|\mathbf{g}_i - \hat{\mathbf{g}}_i\|_2 = \sqrt{\sum_{j \notin \mathcal{I}_i} g_{i,j}^2}.
\label{eq:topk_error}
\end{equation}
This error reflects the amount of information lost due to compression and is typically nonzero unless the original gradient is already sparse. In addition to the magnitude error, Top-K compression also alters the direction of the gradient vector.

\textbf{Comparison with Uncompressed SGD}:

In uncompressed SGD, the entire gradient \(\mathbf{g}_i\) is transmitted without compression. Therefore, the cosine similarity between the reconstructed gradient and the true gradient can theoretically reach 1, allowing the adversary to fully reconstruct the original input \(x\). Specifically, when the adversary minimizes the loss \(\mathcal{L}_{\text{recon}}(x)\), there is no compression-induced error term, implying that:

\begin{equation}
\mathcal{L}_{\text{recon}}(x) \to 0
\end{equation}

\textbf{Implications}:

Due to the divergence in cosine similarity caused by compression, even if the adversary minimizes \(\mathcal{L}_{\text{recon}}(x)\), the best possible alignment between the reconstructed gradient and the true gradient is limited by the compression error. The discarded high-rank information in PowerSGD or the discarded gradient components in Top-K SGD prevent full recovery of the true gradient, thereby enhancing privacy protection. Of course, as the degree of compression decreases, more information is retained, which increases the possibility of successful gradient reconstruction and thereby reduces privacy protection.

\section{Empirical Study}
\label{sec:Experiment}

In this section, we present the empirical analysis of the GradInv on different SGD algorithms, including PowerSGD, Top-K SGD, and the original SGD. We include the experimental setup, attack results, and a discussion of the findings.

\subsection{Experiments Setup} 
The experiments are conducted using the ResNet18 architecture trained on the three baseline datasets: CIFAR-10, CIFAR-100, and MNIST dataset, a widely-used benchmark for image classification tasks, each configuration evaluates 10 reconstructed image samples. We evaluated the vulnerability of different compression strategies under GradInv by comparing the Structural Similarity Index Measure (SSIM) values for the reconstructed data, ranges from 0 to 1, where a value of 1 indicates perfect similarity between two images, and 0 indicates no similarity. The algorithms compared in our analysis included:

\begin{itemize}
    \item PowerSGD: Evaluated at ranks $1, 2, 4, 30,  
       50$.

    \item Top-K SGD: With compression ratios equivalent to the corresponding ranks of PowerSGD.

    \item Original SGD: As a baseline, without any compression.
\end{itemize}

\begin{figure*}[htbp]

\caption{Gradient Inversion Images Across Datesets}
    \centering
    \label{fig:GIA_3datesets}
    
 \makebox[\textwidth][c]{%

        \begin{tabular}{c l l l}
            \hspace*{-1cm} 
            \begin{minipage}[b]{0.15\linewidth}
           
                \centering
                \scriptsize{\textbf{Original Image}}\\[0.2cm]
                \scriptsize{\textbf{SGD}}\\[0.2cm] 
                \scriptsize{\textbf{PowerSGD Rank 1}}\\[0.275cm]
                \scriptsize{\textbf{Top-K SGD Level 1\textsuperscript{*}}}\\[0.275cm]
                \scriptsize{\textbf{PowerSGD Rank 2}}\\[0.275cm]
                \scriptsize{\textbf{Top-K SGD Level 2\textsuperscript{*}}}\\[0.275cm]
                \scriptsize{\textbf{PowerSGD Rank 4}}\\[0.275cm]                \scriptsize{\textbf{Top-K SGD Level 4\textsuperscript{*}}}\\[0.275cm]
                \scriptsize{\textbf{PowerSGD Rank 30}}\\[0.275cm]
                \scriptsize{\textbf{Top-K SGD Level 30\textsuperscript{*}}}\\[0.275cm]
                \scriptsize{\textbf{PowerSGD Rank 50}}\\[0.275cm]
                \scriptsize{\textbf{Top-K SGD Level 50\textsuperscript{*}}}\\[0.275cm]
                \scriptsize{\textbf{}}\\[0.8cm]
            \end{minipage} &
            
            \hspace*{-5mm}
            \begin{minipage}[b]{0.26\linewidth}
                \centering
                \includegraphics[height=0.48cm]{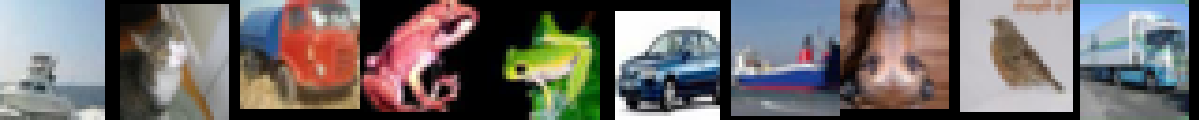}\\[0.06cm]
                \includegraphics[height=0.48cm]{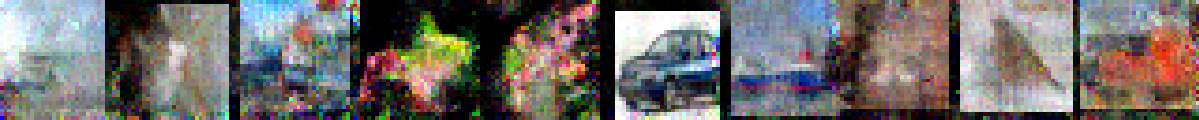}\\[0.06cm]
                \includegraphics[height=0.48cm]{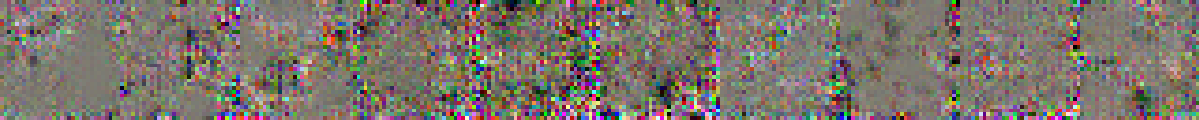}\\[0.06cm]
                \includegraphics[height=0.48cm]{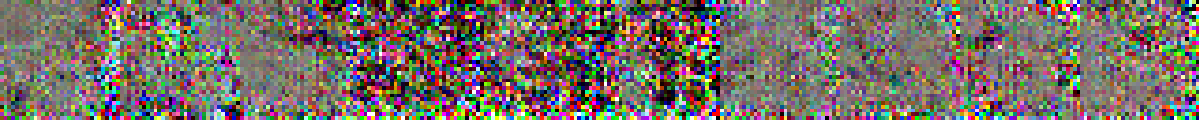}\\[0.06cm]
                \includegraphics[height=0.48cm]{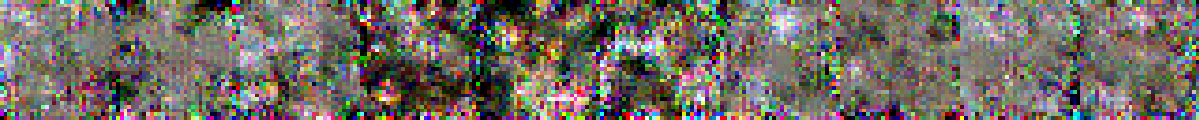}\\[0.06cm]
                \includegraphics[height=0.48cm]{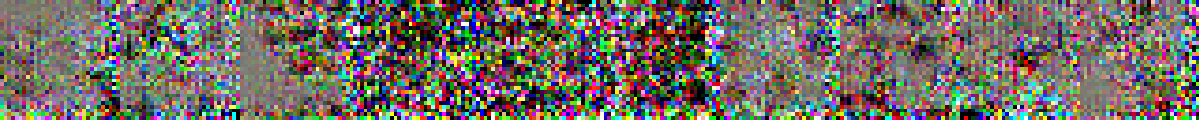}\\[0.06cm]
                \includegraphics[height=0.48cm]{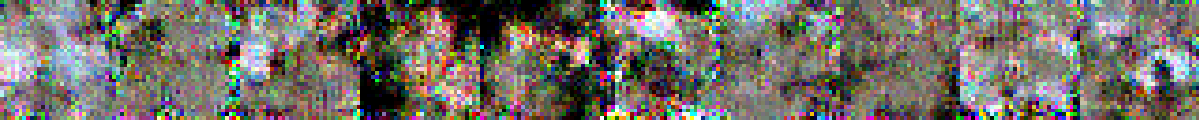}\\[0.06cm]
                \includegraphics[height=0.48cm]{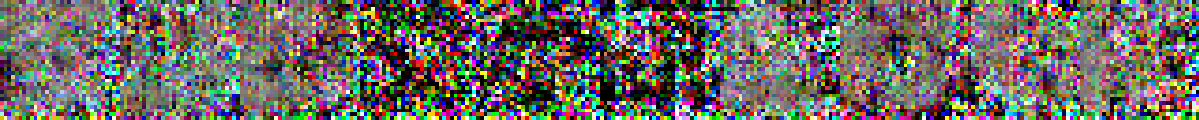}\\[0.06cm]
                \includegraphics[height=0.48cm]{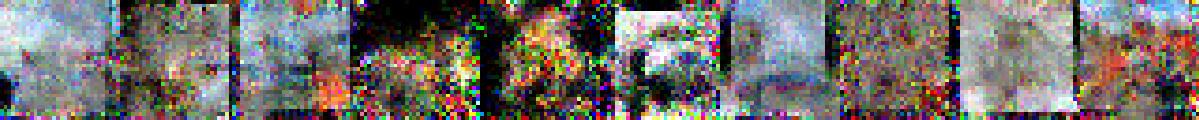}\\[0.06cm]
                \includegraphics[height=0.48cm]{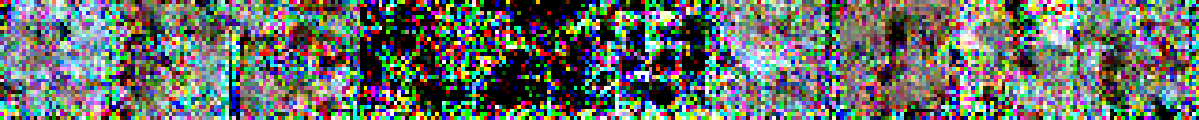}\\[0.06cm]
                \includegraphics[height=0.48cm]{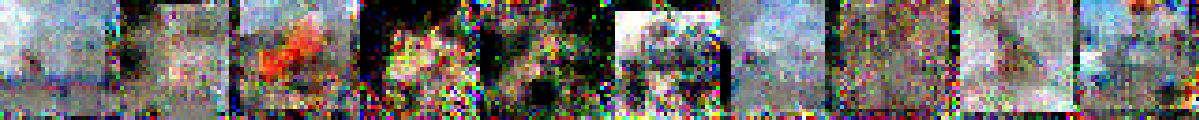}\\[0.06cm]
                \includegraphics[height=0.48cm]{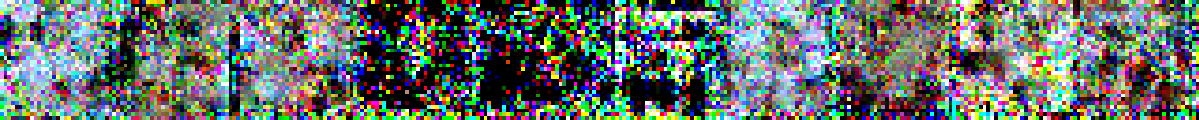}
                \text{(a) CIFAR-10}
            \end{minipage} &

            \begin{minipage}[b]{0.26\linewidth}
                \centering
                \includegraphics[height=0.48cm]{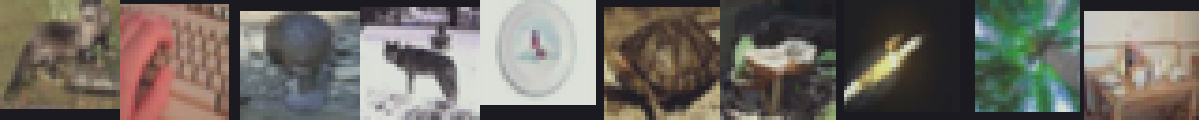}\\[0.06cm]
                \includegraphics[height=0.48cm]{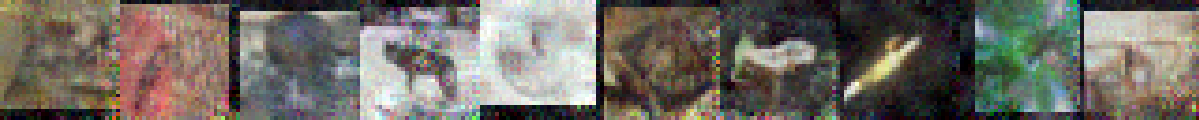}\\[0.06cm]
                \includegraphics[height=0.48cm]{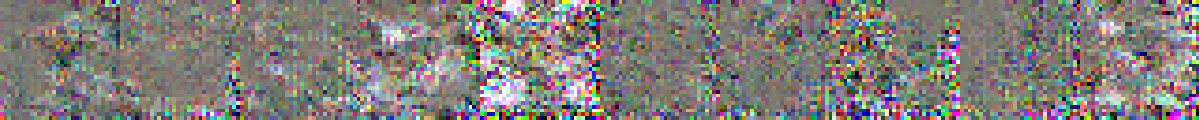}\\[0.06cm]
                \includegraphics[height=0.48cm]{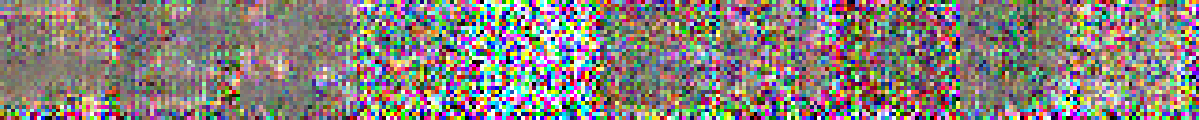}\\[0.06cm]
                \includegraphics[height=0.48cm]{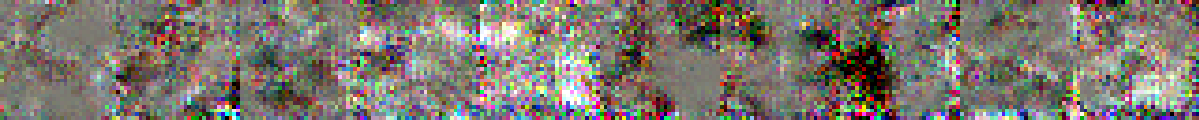}\\[0.06cm]
                \includegraphics[height=0.48cm]{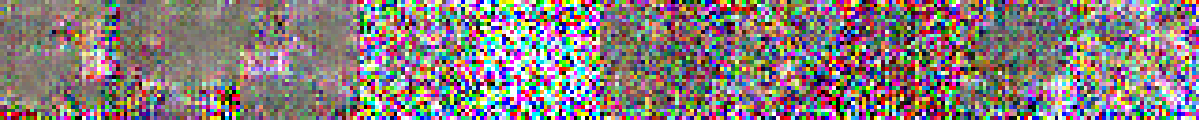}\\[0.06cm]
                \includegraphics[height=0.48cm]{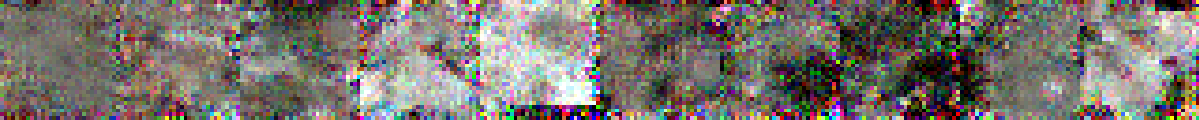}\\[0.06cm]
                \includegraphics[height=0.48cm]{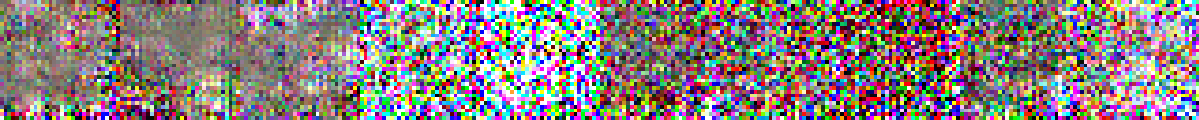}\\[0.06cm]
                \includegraphics[height=0.48cm]{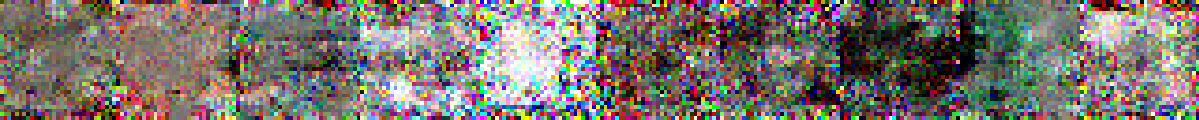}\\[0.06cm]
                \includegraphics[height=0.48cm]{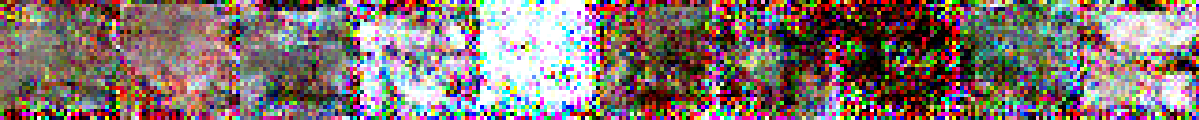}\\[0.06cm]
                \includegraphics[height=0.48cm]{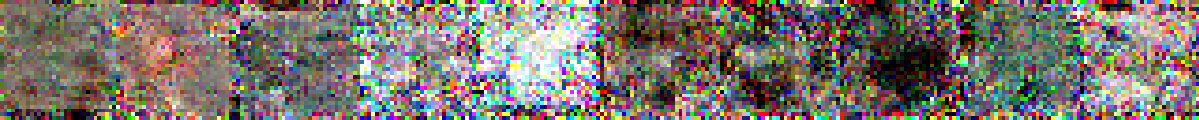}\\[0.06cm]
                \includegraphics[height=0.48cm]{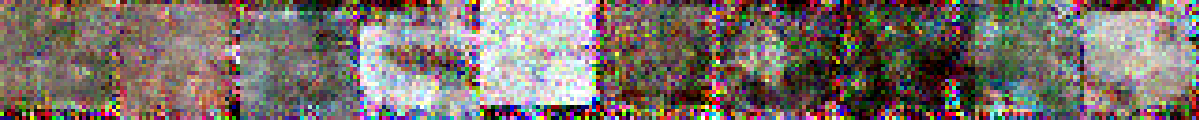}
                \text{(b) CIFAR-100}
            \end{minipage} &

            \begin{minipage}[b]{0.26\linewidth}
                \centering
                \includegraphics[height=0.48cm]{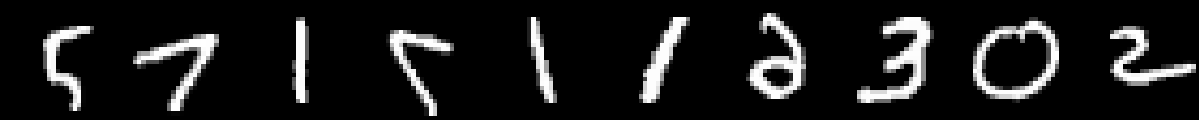}\\[0.06cm]
                \includegraphics[height=0.48cm]{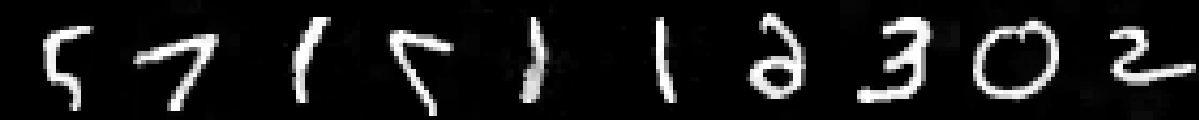}\\[0.06cm]
                \includegraphics[height=0.48cm]{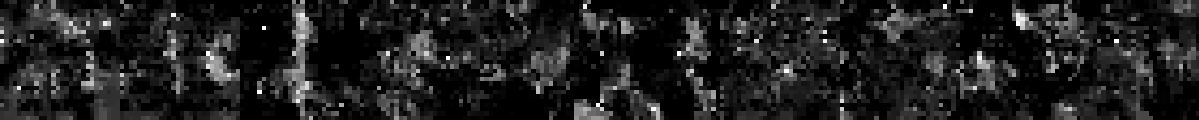}\\[0.06cm]
                \includegraphics[height=0.48cm]{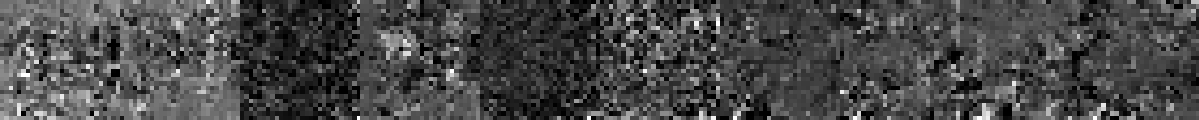}\\[0.06cm]
                \includegraphics[height=0.48cm]{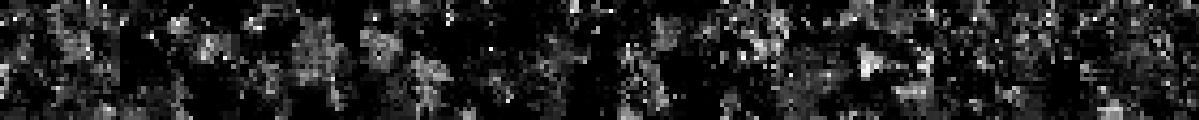}\\[0.06cm]
                \includegraphics[height=0.48cm]{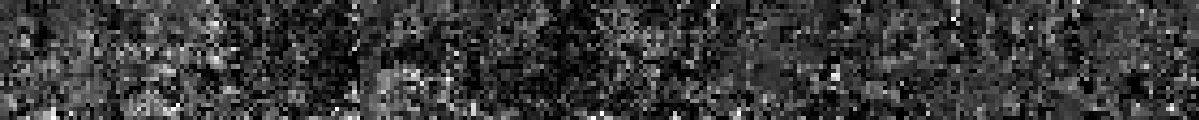}\\[0.06cm]
                \includegraphics[height=0.48cm]{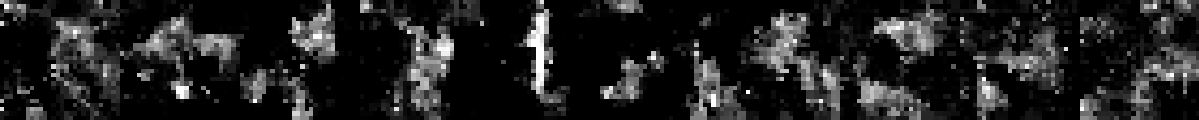}\\[0.06cm]
                \includegraphics[height=0.48cm]{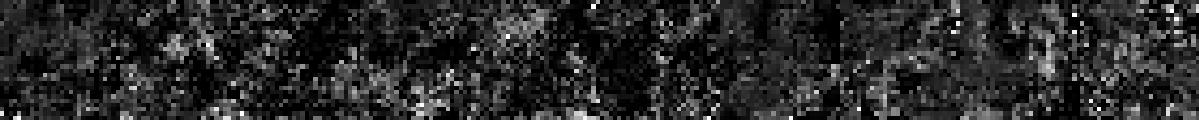}\\[0.06cm]
                \includegraphics[height=0.48cm]{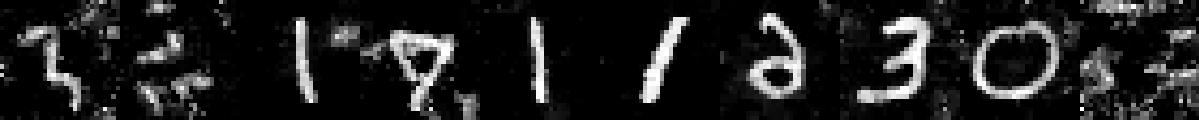}\\[0.06cm]
                \includegraphics[height=0.48cm]{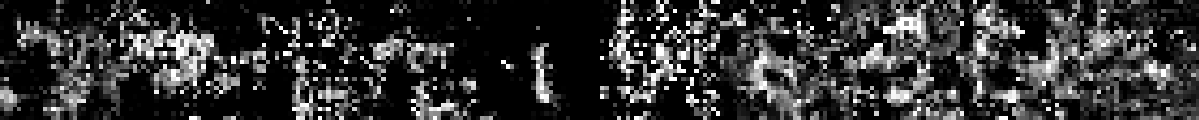}\\[0.06cm]
                \includegraphics[height=0.48cm]{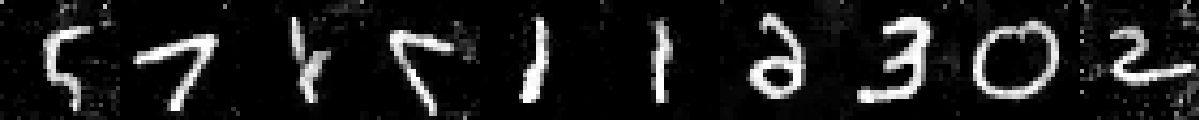}\\[0.06cm]
                \includegraphics[height=0.48cm]{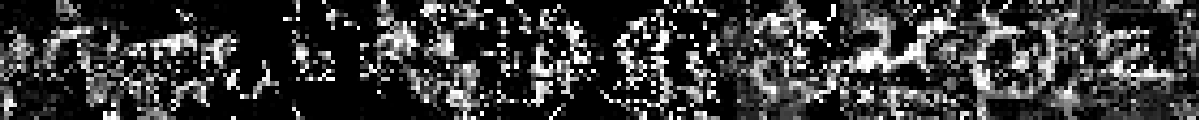}
                \text{(c) MNIST}
            \end{minipage}
        \end{tabular}
}

\end{figure*}


\begin{table*}[ht]
\centering
\caption{SSIM Comparison across Datasets\\\textsuperscript{*}The level of compression rate of Top-K SGD corresponds to the respective rank values used for PowerSGD.}
\label{table.SSIM}

\setlength{\tabcolsep}{3pt}
\begin{tabular}{lccccccccc}
\toprule
Algorithm & \multicolumn{3}{c}{CIFAR-10} & \multicolumn{3}{c}{CIFAR-100} & \multicolumn{3}{c}{MNIST} \\
\cmidrule(r){2-4} \cmidrule(r){5-7} \cmidrule(r){8-10}
 &  Mean &  Std & Percentage & Mean &  Std & Percentage & Mean & Std & Percentage \\
\midrule
Original SGD                             & 0.3451        & $\pm$0.2230  & 100\%   & 0.5576        & $\pm$0.0796  & 100\%   & 0.8685        & $\pm$0.1198  & 100\% \\
PowerSGD (Rank 1)                        & 0.0040        & $\pm$0.0028  & 1.16\%  & 0.0069        & $\pm$0.0056  & 1.24\%  & 0.0708        & $\pm$0.0736  & 8.15\% \\
Top-K SGD (Level 1)\textsuperscript{*}    & 0.0071        & $\pm$0.0051  & 2.06\%  & 0.0069        & $\pm$0.0056  & 1.24\%  & 0.0035        & $\pm$0.0024  & 0.40\% \\
PowerSGD (Rank 2)                        & 0.0137        & $\pm$0.0079  & 3.97\%  & 0.0112        & $\pm$0.0093  & 2.01\%  & 0.0946        & $\pm$0.0659  & 10.89\% \\
Top-K SGD (Level 2)\textsuperscript{*}    & 0.0069        & $\pm$0.0053  & 2.00\%  & 0.0069        & $\pm$0.0049  & 1.24\%  & 0.0072        & $\pm$0.0050  & 0.83\% \\
PowerSGD (Rank 4)                        & 0.0361        & $\pm$0.0306  & 10.46\% & 0.0231        & $\pm$0.0182  & 4.14\%  & 0.2361        & $\pm$0.1519  & 27.19\% \\
Top-K SGD (Level 4)\textsuperscript{*}    & 0.0065        & $\pm$0.0051  & 1.88\%  & 0.0085        & $\pm$0.0059  & 1.52\%  & 0.0156        & $\pm$0.0078  & 1.80\% \\
PowerSGD (Rank 30)                       & 0.1101        & $\pm$0.0758  & 31.91\% & 0.0437        & $\pm$0.0306  & 7.84\%  & 0.4909        & $\pm$0.2228  & 56.53\% \\
Top-K SGD (Level 30)\textsuperscript{*}   & 0.0178        & $\pm$0.0151  & 5.16\%  & 0.0385        & $\pm$0.0306  & 6.91\%  & 0.1706        & $\pm$0.1761  & 19.64\% \\
PowerSGD (Rank 50)                       & 0.1339        & $\pm$0.0542  & 38.81\% & 0.0482        & $\pm$0.0327  & 8.65\%  & 0.6132        & $\pm$0.1155  & 70.63\% \\
Top-K SGD (Level 50)\textsuperscript{*}   & 0.0262        & $\pm$0.0169  & 7.59\%  & 0.0758        & $\pm$0.0401  & 13.59\% & 0.1715        & $\pm$0.1183  & 19.75\% \\

\bottomrule
\end{tabular}
\end{table*}

\subsection{Results and Analysis}
\label{sec:result}

Fig. ~\ref{fig:GIA_3datesets} displays the reconstructed images from GradInv and Table~\ref{table.SSIM} summarizes the SSIM values for reconstructed images, under different algorithms and compression settings across datasets. The metrics reported include the mean SSIM, standard deviation, and percentage of the SSIM relative to the baseline obtained with original SGD (set as 100 \%). The results highlight how each compression method affects the quality of gradient inversion, providing insight into the privacy-preserving capability of each algorithm:

Original SGD: consistently results in the highest SSIM values, indicating high-quality reconstructions, reflecting a higher risk of privacy leakage. In all data sets, the SSIM achieved by Original SGD remains at 100\%, serving as the baseline.

PowerSGD (Rank 1, 2, 4, 30, 50): As the rank increases, the mean SSIM value also increases, which implies that higher ranks lead to more information being retained during compression. For example, PowerSGD with Rank 50 achieves an SSIM of 0.6132 on the MNIST dataset, which is significantly higher than lower ranks. The trend is consistent across all datasets, with larger ranks leading to improved reconstructions but also potentially increased privacy risks. Top-K SGD (Level 1, 2, 4, 30, 50): The Top-K approach generally shows lower SSIM values compared to PowerSGD at equivalent compression rates. For instance, Top-K SGD at level 50 achieves an SSIM of 0.1715 for MNIST, which is notably lower than PowerSGD at the same compressed ratio. This suggests that Top-K SGD might be more effective at reducing privacy leakage compared to PowerSGD.
\textbf{Dataset-Specific Analysis}: MNIST, being a simpler dataset, consistently shows higher SSIM values across all algorithms compared to CIFAR-10 and CIFAR-100, suggesting that gradient inversion attacks are more effective on datasets with simpler data distributions. For CIFAR-10 and CIFAR-100, the SSIM values are generally lower, indicating better resistance to GradInv under compression methods.

\begin{table*}[ht]
\caption{MIAs across datasets \\\textsuperscript{*}The level of compression rate of Top-K SGD corresponds to the respective rank values used for PowerSGD.}
\centering
\begin{tabular}{l@{\hspace{1cm}}l@{\hspace{1cm}} c@{\hspace{1cm}} c@{\hspace{1cm}} c@{\hspace{1cm}}}
\hline
\textbf{Datasets} & \textbf{Algorithms} & \textbf{Prediction MIA} & \textbf{Loss MIA} & \textbf{Cross-Entropy MIA} \\
\hline
\multirow{11}{*}{\textbf{CIFAR-10}} 
 & OriginalSGD & 54.02\% & 55.17\% & 55.93\% \\
 & PowerSGD (Rank 1) & 54.10\% & 55.15\% & 56.05\% \\
 & Top-K SGD (Level 1)\textsuperscript{*} & 54.93\% & 54.87\% & 53.03\% \\
 & PowerSGD (Rank 2)\textsuperscript{*} & 53.98\% & 54.92\% & 56.47\% \\
 & Top-K SGD (Level 2) & 56.20\% & 56.30\% & 54.97\% \\
 & PowerSGD (Rank 4) & 54.02\% & 55.07\% & 55.88\% \\
 & Top-K SGD (Level 4)\textsuperscript{*} & 56.52\% & 57.20\% & 56.53\% \\
 & PowerSGD (Rank 30) & 54.23\% & 55.08\% & 56.03\% \\
 & Top-K SGD (Level 30)\textsuperscript{*} & 54.15\% & 55.08\% & 56.00\% \\
 & PowerSGD (Rank 50) & 53.83\% & 54.80\% & 55.53\% \\
 & Top-K SGD (Level 50)\textsuperscript{*} & 54.13\% & 55.12\% & 55.77\% \\
\hline
\multirow{11}{*}{\textbf{CIFAR-100}} 

 & OriginalSGD & 66.78\% & 69.83\% & 72.57\% \\
 & PowerSGD (Rank 1) & 67.27\% & 70.23\% & 71.52\% \\
 & Top-K SGD (Level 1)\textsuperscript{*} & 56.25\% & 55.73\% & 51.68\% \\
 & PowerSGD (Rank 2) & 66.70\% & 70.08\% & 71.73\% \\
 & Top-K SGD (Level 2)\textsuperscript{*} & 57.35\% & 56.90\% & 52.47\% \\
 & PowerSGD (Rank 4) & 66.82\% & 70.17\% & 72.50\% \\
 & Top-K SGD (Level 4)\textsuperscript{*} & 63.47\% & 63.03\% & 56.18\% \\
 & PowerSGD (Rank 30) & 67.27\% & 70.17\% & 72.47\% \\
 & Top-K SGD (Level 30)\textsuperscript{*} & 67.82\% & 69.85\% & 69.45\% \\
 & PowerSGD (Rank 50) & 67.35\% & 70.27\% & 72.67\% \\
 & Top-K SGD (Level 50)\textsuperscript{*} & 67.20\% & 69.47\% & 70.25\% \\
\hline
\multirow{11}{*}{\textbf{MNIST}} 
 & OriginalSGD & 51.00\% & 51.30\% & 51.48\% \\
 & PowerSGD (Rank 1) & 51.18\% & 51.57\% & 51.80\% \\
 & Top-K SGD (Level 1)\textsuperscript{*} & 51.17\% & 51.60\% & 51.77\% \\
 & PowerSGD (Rank 2) & 51.22\% & 51.62\% & 51.85\% \\
 & Top-K SGD (Level 2)\textsuperscript{*} & 51.22\% & 51.77\% & 51.83\% \\
 & PowerSGD (Rank 4) & 51.05\% & 51.48\% & 51.80\% \\
 & Top-K SGD (Level 4)\textsuperscript{*} & 51.08\% & 51.53\% & 51.73\% \\
 & PowerSGD (Rank 30) & 51.08\% & 51.48\% & 51.68\% \\
 & Top-K SGD (Level 30)\textsuperscript{*} & 51.03\% & 51.38\% & 52.02\% \\
 & PowerSGD (Rank 50) & 50.98\% & 51.40\% & 51.62\% \\
 & Top-K SGD (Level 50)\textsuperscript{*} & 50.97\% & 51.35\% & 51.53\% \\
\hline
\end{tabular}
\label{table:mia}
\end{table*}

\section{Discussion}

Following the setup detailed in Section~\ref{sec:Experiment}, we conducted membership inference attacks (MIA) of three types---prediction-based, loss-based, and cross-entropy-based---on multiple datasets (CIFAR-10, CIFAR-100, MNIST) and gradient compression algorithms (OriginalSGD, PowerSGD, Top-K SGD). The results, summarized in Table~\ref{table:mia}, present the attack success rates under various settings.

Contrary to our initial hypothesis, compressing gradients did not generally lead to a substantial reduction in MIA success rates. Across CIFAR-10, for example, the differences between OriginalSGD and compressed variants (PowerSGD, Top-K SGD) are minimal, suggesting that the compressed gradients still retain sufficient information for adversaries. Likewise, on MNIST, all tested compression methods yield almost the same attack success rates as the uncompressed baseline.

However, in certain configurations—particularly for CIFAR-100—Top-K SGD with higher compression levels (e.g., Level~1 and Level~2) shows a more pronounced decrease in attack success. This indicates that specific compression strategies at certain intensities can have a tangible impact on MIA resistance, though this pattern is not universal across all datasets and algorithms.

Overall, these observations highlight a more complex relationship between gradient compression and MIA success than initially anticipated. On the one hand, gradient compression may preserve enough information for adversaries to exploit. On the other hand, current black-box MIA techniques may be insufficiently sensitive to subtle privacy leakages in machine learning, especially in distributed learning. This aligns with recent findings suggesting that existing MIA metrics can be misleading in assessing privacy risks~\cite{li2024privacy, aerni2024evaluations}.

\section{Conclusion}
\label{sec:conclusion}

This study examines the impact of gradient compression techniques, including PowerSGD and Top-K SGD, and uncompressed SGD on the effectiveness of GradInv and MIA in distributed learning across various datasets.  We provide comprehensive empirical evaluations to assess the privacy risks of different SGD variants, addressing gaps in the understanding of these methods’ trustworthiness.

Our findings show that uncompressed gradients using Original SGD pose the highest privacy risk, as demonstrated by high SSIM values and detailed reconstructions. In contrast, gradient compression techniques significantly lower the quality of reconstructed images, thereby enhancing privacy protection. Notably, we find that MIA demonstrates low sensitivity to both compressed and uncompressed SGD, indicating that MIA may not be a reliable metric for assessing privacy risks in distributed machine learning. We call for further investigation into the existing privacy evaluation methods for machine learning. 

Future work will explore the effects of compression on other attack methods, such as data poisoning and backdoor attacks, to gain a more complete understanding of the privacy implications of these methods. 

\bibliographystyle{IEEEbib}
\bibliography{main}

\end{document}